\documentclass{article}

\usepackage[utf8]{inputenc} %
\usepackage[T1]{fontenc}    %
\usepackage{hyperref}       %
\usepackage{url}            %
\usepackage{booktabs}       %
\usepackage{amsfonts}       %
\usepackage{nicefrac}       %
\usepackage{microtype}      %
\usepackage{xcolor}         %

\usepackage{amsfonts,amssymb,amsmath}
\usepackage{graphicx}
\usepackage{paralist}
\usepackage{xcolor} %
\usepackage{natbib}

\renewcommand{\emptyset}{{\varnothing}}

\newcommand{\giv}{\!\mid\!}

\newcommand{\real}{\mathbb{R}}

\newcommand{\bsw}{\boldsymbol{w}}
\newcommand{\bsx}{\boldsymbol{x}}

\newcommand{\bsz}{\boldsymbol{z}}

\newcommand{\val}{\mathrm{val}}

\newcommand{\e}{\mathbb{E}}
\newcommand{\var}{\mathrm{var}}

\newcommand{\pr}{\mathrm{Pr}}

\newcommand{\olt}{\overline\tau}
\newcommand{\ult}{\underline\tau}

\newcommand{\phm}{\phantom{-}}

\newcommand{\cx}{\mathcal{X}}

\newcommand{\fp}{\mathrm{FP}}
\newcommand{\fn}{\mathrm{FN}}
\newcommand{\sumdot}{\text{\tiny$\bullet$}}

\newcommand{\fpr}{\mathrm{FPR}}
\newcommand{\fnr}{\mathrm{FNR}}
\newcommand{\tpr}{\mathrm{TPR}}
\newcommand{\ppv}{\mathrm{PPV}}

\title{Cohort Shapley value for algorithmic fairness}

\author{
   Masayoshi Mase\\Hitachi, Ltd.
   \and
   Art B. Owen \\ Stanford University
   \and
   Benjamin B. Seiler \\ Stanford University
}
\date{May 2021}

\begin{document}

\maketitle

\begin{abstract}
  Cohort Shapley value is a model-free method of
  variable importance grounded in game theory that
  does not use any unobserved and potentially
  impossible feature combinations.
  We use it to evaluate algorithmic fairness, using
  the well known COMPAS recidivism
  data as our example.
  This approach allows one to identify
  for each individual in a data set the extent to
  which they were adversely or beneficially
  affected by their value of a
  protected attribute such as their race.
  The method can do this even if race was not
  one of the original predictors and even if
  it does not have access to a proprietary
  algorithm that has made the predictions.
  The grounding in game theory lets us define aggregate
  variable importance for a data set consistently
  with its per subject definitions.
  We can   investigate variable importance for multiple
 quantities of  interest in the fairness literature
 including false positive predictions.
\end{abstract}

\section{Introduction}

Machine learning is now commonly used to make consequential decisions about people, affecting
hiring decisions, loan applications,
medical treatments,
criminal sentencing and more.
It is important to understand and explain these decisions.
A critical part of understanding a decision is 
quantifying the importance of the variables 
used to make it.  When the decisions
are about people and some of the variables describe
protected attributes of those people, such as race and
gender, then variable importance has a direct bearing
on algorithmic fairness.

As we describe below, most variable importance
measures work by changing a subset of
the input variables to a black box.  We survey
such methods below and argue against that approach.
The first problem is that the resulting analysis
can depend on some very unreasonable variable 
combinations.  A second problem with changing inputs
to a black box is that it necessarily attributes zero
importance to a variable that the algorithm did not
use.  If a protected variable is not actually used in
the black box then the algorithm would automatically
be considered fair.  However this `fairness through
unawareness' approach is not reliable,
as \cite{adle:2018} and many others have noted.
Information about the protected variables can leak
in through others with which they are associated.
The practice known as `redlining' involves
deliberate exploitation of such associations.
When studying fairness we must have the possibility
of studying a variable not included in the black
box.  In this paper we develop and illustrate
an approach to algorithmic fairness that does not
use impossible values and can detect redlining.

\subsection{Variable Importance}

Variable importance measures have a long
history and there has been a recent surge in
interest motivated by problems of
explainable AI.
The global sensitivity analysis literature studies black box functions
used in engineering and climate models among others.
Much of that work is based
on a functional ANOVA model of the function relating
the output to inputs. Variance explained
is partitioned using Sobol' indices
\citep{sobo:1993}.
\cite{salt:etal:2008} is an introductory
textbook and \cite{raza:etal:2021} is a current survey
of the field.  \cite{wei:lu:song:2015} provide
a comprehensive survey of variable importance measures in statistics. They
include 197 references of which 24 are
themselves surveys.
\cite{moln:2018} surveys variable importance
measures used in explainable AI.
Prominent among these are SHAP \citep{lund:lee:2017}
and LIME \citep{ribe:etal:2016}.  LIME makes a local linear
approximation to a black box function $f(\cdot)$
and one can then take advantage of the
easier interpretability of linear models.
SHAP is based on Shapley value from
cooperative game theory \citep{shap:1953}, that
we describe in more detail below.

Ordinarily we can compare two prediction methods
by waiting to see how accurate they are on unknown
future data or, if necessary, use
holdout data.
This does not carry over to deciding
whether LIME
or SHAP or some other method has made a
better explanation of a past decision.
Suppose that we want to understand something like why a
given applicant was turned down for a loan.
When $f(\cdot)$ is available to us in closed form
then there is no doubt at all about whether the
loan would have been offered under any assortment
of hypothetical feature combinations that we choose.
We can compare methods by
how they define importance.  Choosing among
definitions is a different activity, essentially
a philosophical one, and we face tradeoffs.
Our problem is one of identifying the causes
of given effects when we have perfect knowledge 
of $f(\cdot)$ for all input variable
combinations.  This is different from the more
common learning task of quantifying the effects
of given causes. That distinction was made by
\cite{holl:1988} in studying causal inference. It
goes back at least to \cite{mill:1843}.

A problem with many, but not all, measures of
variable importance is that they are based
on changing some feature values from one level
to another, while holding other features constant
and then looking at the changes to $f(\cdot)$.
When the underlying features are
highly correlated, then some of these combinations
can be quite implausible, casting doubt on any
variable importance methods that use them.
In extreme cases, the combinations can be physically
impossible (e.g., systolic blood pressure below
diastolic) or logically impossible (e.g., birth date
after graduation date).  The use of these combinations
has also been criticized by \cite{hook:ment:2019:tr}.

To avoid using impossible values, Cohort Shapley method\footnote{https://github.com/cohortshapley/cohortshapley}
from \cite{mase:owen:seil:2019} uses only observed data values.
For any target subject $t\in\{1,\dots,n\}$ and any
feature variable $j=1,\dots,d$ we obtain
a Shapley value $\phi_j=\phi_j(t)$ that measures
the impact of the value of feature $j$ on $f(\cdot)$
for subject $t$.  As we describe below, those
impacts can be positive or negative.  Using one of
the Shapley axioms we will be able to aggregate from
individual subjects to an impact measure for the
entire data set. We can also
disaggregate from the entire data set to a subset,
such as all subjects with the protected level of a
protected variable.

\subsection{Contributions}

\begin{figure}
    \centering
    \includegraphics{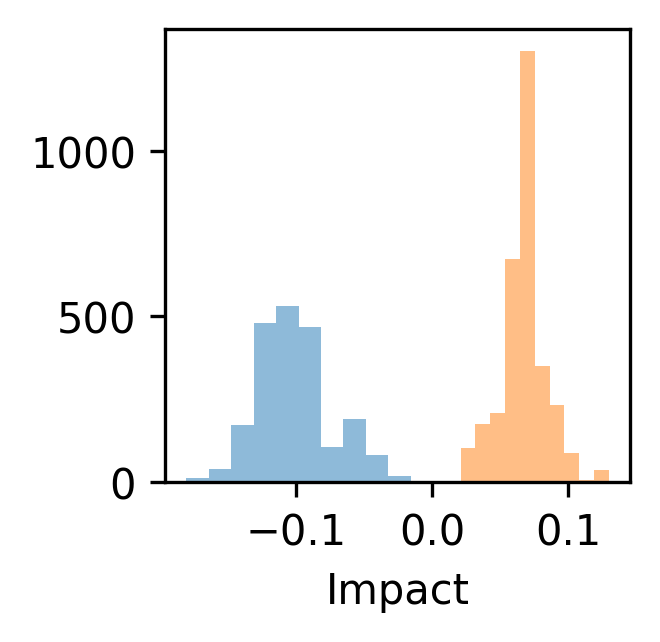}
    \caption{
Histograms of the cohort Shapley
impact of race on whether a person in the COMPAS
data is predicted to reoffend. Orange bars represent
Black subjects; blue represent White subjects.
    }
    \label{fig:compas_dp_cs_r_hist}
\end{figure}

In this paper we use the cohort Shapley method
from \cite{mase:owen:seil:2019} to measure
variable importance.
One of the features of cohort Shapley is
that it can attribute importance to a variable
that is not actually used in $f(\cdot)$.
This is controversial and not all authors
approve but it is essential in the present
context of algorithmic fairness.

Our contribution is to provide a method of quantifying bias at 
both individual and group levels with axiomatic consistency between 
the individual and group measures and without requiring any impossible 
or unobserved  combinations of input variables.
The approach can even
be used when the prediction algorithm 
is not available as for instance when it is proprietary.

We investigate the COMPAS data \citep{angw:2016}
which include predictions of who is likely
to commit a crime.  There is great interest
in seeing whether the algorithm is unfair
to Black subjects.  In the given context,
the prediction code $f(\cdot)$ is a proprietary
algorithm, unavailable to us and so we cannot
change any of the values in it. What we have
instead are the predictions on a set of subjects.
Cohort Shapley can work with the predictions
because it does not require any
hypothetical feature combinations.
Figure~\ref{fig:compas_dp_cs_r_hist} shows cohort
Shapley impacts for the race of the subjects
in the COMPAS data, computed in a way that we
describe below.  As it turns out the impact
of race on whether a subject was predicted to
reoffend was always positive for Black subjects
and it was always
negative for White subjects. For some other
measures that we show, the histograms
overlap.  The average impact for
Black subjects was $0.067$ and the average
for White suspects was $-0.101$.
For context, the response values were $1$ for
those predicted to reoffend and $0$ for those
predicted not to reoffend.

An outline of this paper is as follows.
Section~\ref{sec:basics} introduces
our notation, defines Shapley value,
and presents a few of the variable
importance measures and fairness definitions from the literature.
Section~\ref{sec:compas} describes
the COMPAS recidivism data set.
Section~\ref{sec:dataanalysis} has
our analysis of that data including
a Bayesian bootstrap for uncertainty quantification.
Section~\ref{sec:conclusions} has
our conclusions.
An appendix presents some additional COMPAS
results like the ones we selected for
discussion.

\section{Notation and definitions}\label{sec:basics}

For subjects $i=1,\dots,n$, the value of feature
$j=1,\dots,d$ is $x_{ij}\in\cx_j$.
We consider categorical variables $x_{ij}$.
Continuously distributed values can
be handled as discussed in
Section~\ref{sec:conclusions}.
The features for subject $i$ are
encoded in $\bsx_i\in\cx =\prod_{j=1}^d\cx_j$.
For each subject there is a response value
$y_i\in \real$.
The feature indices belong to the set $1{:}d\equiv\{1,\dots,d\}$ and similarly
the subject indices are in $1{:}n\equiv\{1,\dots,n\}$.
For $u\subseteq1{:}d$ the tuple $\bsx_u$
is $(x_j)_{j\in u}\in\cx_u=\prod_{j\in u}\cx_j$,
and $\bsx_{iu}=(x_{ij})_{j\in u}$.

There is also an algorithmic
prediction $\hat y_i$.
It is usual in variable importance problems
to have $\hat y_i =f(\bsx_i)$ for a function $f(\cdot)$
that may be difficult to interpret (e.g., a black box).
This $f(\cdot)$ is often an approximation of
$\e(y\giv \bsx)$ though it need not be of that form.
Cohort Shapley uses only the values $\hat y_i$ so it
does not require access to $f(\cdot)$.  Since it
only uses a vector of values, one per subject, it
can be used to find the important variables in $y_i$,
or $\hat y_i$ or combinations such as
the residual $y_i-\hat y_i$.

In many settings $y_i$ and $\hat y_i$ are both
binary. The value $y_i=1$ may mean that subject $i$
is worthy of a loan, or is predicted to commit a crime,
or should be sent to intensive care and $\hat y_i\in\{0,1\}$
is an estimate of $y_i$.  In this case,
measures such as $\fp_i = 1\{\hat y_i=1\ \&\ y_i=0\}$
describing a false positive are of interest.

Here are some typographic conveniences
that we use.
When $j\not\in u$, then $u+j$
is $u\cup\{j\}$. For $u\subseteq1{:}d$
we use $-u$ for $1{:}d\setminus u$.
The subscript to $\bsx$ may contain
a comma or not depending on what is
clearer. For instance, we use $\bsx_{iu}$
but also $\bsx_{i,-u}$.

\subsection{Shapley values}
Shapley value \citep{shap:1953} is used in game theory
to define a fair allocation of rewards to a team that
has cooperated to produce something of value.
Many variable importance problems can be formulated
as a team of input variables generating an output
value or an output variance explained. We then
want to apportion importance or impact to the
individual variables.

Suppose that a team of $d$ members produce
a value $\val( 1{:}d)$, and that we have at our disposal
the value $\val(u)$ that would have been produced
by the team $u$,
for all $2^d$ teams $u\subseteq1{:}d$.
Let $\phi_j$ be the reward for player $j$.
It is convenient to work with incremental
values $\val(j\giv u)=\val(u+j)-\val(u)$
for sets $u$ with $j\not\in u$.

Shapley introduced quite reasonable criteria:
\begin{compactenum}[\quad 1)]
\item Efficiency: $\sum_{j=1}^d\phi_j = \val(1{:}d)$.
\item Symmetry: If $\val(i\giv u)=\val(j\giv u)$ for all $u\subseteq1{:}d\setminus\{i,j\}$,
then $\phi_i=\phi_j$.
\item Dummy: if $\val(j\giv u) = 0$ for all $u\subseteq1{:}d\setminus\{j\}$,
then $\phi_j=0$.
\item Additivity:
if $\val(u)$ and $\val'(u)$ lead to values $\phi_j$ and $\phi_j'$
then the game producing $(\val+\val')(u)$ has values $\phi_j+\phi'_j$.
\end{compactenum}
He found that the unique valuation that satisfies all
four of these criteria is
\begin{align}\label{eq:shapj}
\phi_j = \frac1d\sum_{u\subseteq -j}
{d-1\choose |u|}^{-1}\val(j\giv u)
\end{align}

Formula~\eqref{eq:shapj} is not very intuitive.
Another way to explain Shapley value is as follows.
We could build a team from $\emptyset$
to $1{:}d$ in $d$ steps, adding one member at a time.
There are $d!$ different orders in which
to add team members. The Shapley value $\phi_j$
is the increase in value coming from the addition of
member $j$, averaged over all $d!$ different orders.
From equation~\eqref{eq:shapj} we see that Shapley value does not change if we add or  subtract the same quantity from all $\val(u)$. It can be convenient to make $\val(\emptyset)=0$.

We discuss some variable importance measures
based on Shapley value below.
We use the framework from \cite{sund:najm:2019}.
They survey many
uses of Shapley value in explainable AI.
\subsection{Changing variables}

Most mechanisms for investigating variable
importance proceed by changing some, but not all,
of the input variables to $f(\cdot)$.
For $u\subseteq1{:}d$ the hybrid
point $\bsw=\bsx_{u}{:}\bsz_{-u}$ has
$$
w_j = \begin{cases}
x_{j}, & j\in u\\
z_{j}, & j\not\in u.
\end{cases}
$$
We can investigate changes to $\bsx_{u}$
by examining $f(\bsx_{u}{:}\bsz_{-u})-f(\bsx)$
for various points $\bsx,\bsz\in\cx$.

Global sensitivity analysis \citep{raza:etal:2021} works with
an analysis of variance (ANOVA) defined
in terms of $\bsx\in\cx$ with independent
components for which
$\sigma^2 = \var( f(\bsx))<\infty$.
The sets $\cx_j$ can be discrete or
continuous. It is common there to measure
the importance of a subset of variables by
variance explained.
For instance,  the lower Sobol' index
for variables $\bsx_u$ is
$\ult^2_u=\var( \e(f(\bsx)\giv\bsx_u))$
which is usually normalized to $\ult^2_u/\sigma^2$.
The upper Sobol' index is $\olt^2_u=\sigma^2-\ult^2_{-u}$,
that is, everything not explained by $\bsx_{-u}$
is attributed to $\bsx_u$.
These measures are very natural in settings
where the components of $\bsx$ can all
vary freely and independently,
but such is usually not the
case for inputs to a black box machine
learning model.

The quantitative influence function (QII)
method of \cite{datt:2016}
uses a model in which the features
are statistically independent from
their marginal distributions.

In baseline Shapley (see \cite{sund:najm:2019}),
there is a baseline tuple $\bsx_b\in\cx$
of $\bsx$, such as the sample average
$\bar\bsx$. Then the value for
variable subset $u$ and subject $t$
is $f(\bsx_{t,u}{:}\bsx_{b,-u})$.
We get the same Shapley values using
$\val(u)=f(\bsx_{t,u}{:}\bsx_{b,-u})-f(\bsx_b)$.
Then the total value to explain for
subject $t$ is $f(\bsx_t)-f(\bsx_b)$.
The counterfactuals are changes
in some subset of the values in $\bsx_t$.
Important variables are those that move
the value the most from the baseline
$f(\bsx_b)$ to subject $t$'s value.
Shapley's combination provides a principled
weighting of the effect of changing
$x_{b,j}$ to $x_{t,j}$ given some subset $u$
of variables that have also been changed.

Conditional expectation Shapley
\citep{sund:najm:2019} uses a joint
distribution $D$ on $\cx$ and then
for subject $t$
$\val(u)=\e_D(f(\bsx)\mid\bsx_u=\bsx_{t,u})$
under $D$.
SHAP \citep{lund:lee:2017} is a conditional
expectation Shapley using
a distribution $D$ in which the
features are independent.
Cohort Shapley \citep{mase:owen:seil:2019} that we
describe next is very nearly
conditional expectation Shapley
with $D$ equal to the empirical
distribution.

\subsection{Changing knowledge}

In cohort Shapley the counterfactual
does not involve replacing
$\bsx_{tu}$ by $\bsx_{bu}$. It is instead
about concealing the values of $\bsx_{tu}$.
It is what \cite{kuma:2020} call a
conditional
method because it requires specification
of a conditional distribution on
the features.
The version we present here is a form
of conditional expectation Shapley
using the empirical distribution of
the data.

For every pair of values $x_{tj},x_{ij}\in\cx_j$
we can declare them to be similar or not.
In the present context we take
similarity to just be $x_{tj}=x_{ij}$.
For a target subject $t\in1{:}n$ and
set $u\subseteq1{:}d$ we define
the cohort
$C_u = C_{tu} =\{i\in1{:}n\mid \bsx_{iu}=\bsx_{tu}\}$.
These are the subjects who match subject
$t$ on all of the variables $j\in u$.
They may or may not match the target
on $j\not\in u$.
None of the cohorts is empty because
they all contain $t$.
The value function in cohort Shapley is
the cohort mean
$$
\val(u) = \frac1{|C_{tu}|}
\sum_{i\in C_{tu}}\hat y_i.
$$
Here $\hat y_i=f(\bsx_i)$ but we write
it this way because in practice we might
not have $f(\cdot)$ at our disposal,
just the predictions $\hat y_i$.

The incremental value
$\val(j\giv u)$ is then how much
the cohort mean moves when we reveal
$x_{tj}$ in addition to previously
revealed variables $\bsx_{tu}$.
Having specified the values, the Shapley
formula provides attributions.
Note that $\val(\emptyset)$
is the plain average of all $\hat y_i$
and Shapley value is unchanged by
subtracting $\val(\emptyset)$
from all of the $\val(u)$.
Furthermore, when there are many variables,
then very commonly $C_{t,1{:}d}=\{t\}$
and then the total value to be
explained, $\val(1{:}d)-\val(\emptyset)$
is simply $\hat y_t-(1/n)\sum_{i=1}^n\hat y_i$.

\subsection{Definitions of fairness}

Just as there are many ways to define variable
importance, there are multiple ways to define
what fairness means.  Some of those definitions
are mutually incompatible and some of them differ
from legal definitions.  Here we present a few of
the issues.
See \cite{corb:goel:2018}, \cite{chou:roth:2018}, \cite{berk:2018}, and \citet{frie:etal:2019} for surveys.
We do not make assertions about which
definitions are preferable.

For $y,\hat y\in\{0,1\}$, let $n_{y\hat y}$ be
the number of subjects with
$y_i=y$ and $\hat y_i=\hat y$.
These four counts and their derived properties can
be computed for any subset of the subjects.
The false positive rate (FPR) is
$n_{01}/n_{0\sumdot}$, where a bullet indicates that we are summing over the levels of that index.
We ignore uninteresting corner cases such
as $n_{0\sumdot}=0$; when there are no
subjects with $y=0$ then we have no interest
in the proportion of them with $\hat y=1$.
The false negative rate (FNR) is
$n_{10}/n_{1\sumdot}$.
The prevalence of the trait under study
is $p=n_{1\sumdot}/n_{\sumdot\sumdot}$.
The positive predictive value (PPV) is
$n_{11}/n_{\sumdot1}$.
As \citet[equation (2.6)]{chou:2017} notes,
these values satisfy
\begin{align}\label{eq:fprandp}
\fpr = \frac{p}{1-p}\frac{1-\ppv}{\ppv}(1-\fnr).
\end{align}
See also \cite{klei:mull:ragh:2016}.

Equation~\eqref{eq:fprandp} shows how some natural
definitions of fairness conflict.
FPR and FNR describe $\hat y\giv y$,
while PPV describes $y\giv\hat y$. If two
subsets of subjects have the same PPV but
different prevalences $p$, then they cannot
also match up on FPR and FNR. Fairness in
$y\giv \hat y$ terms and fairness in $\hat y\giv y$
terms can only coincide in trivial settings
such as when $\hat y=y$ always or empirically unusual
settings with equal prevalence between
subjects having different values of a protected
attribute.

\cite{hard:2016} writes about measures
of demographic parity, equalized odds and equal opportunity.
Demographic parity, requires
$\Pr( \hat y=1\giv \bsx_p)=\Pr(\hat y=1)$
where $\bsx_p$ specifies the levels of
one (or more) protected
variables.
Equalized odds requires $\Pr(\hat y=1\giv y=y', \bsx_p) 
= \Pr(\hat y=1\giv y=y'), y' \in {0,1}$, so that false 
positive rates and true positive rates are 
both equal across groups. Because $\fnr = 1 - \tpr$ the 
false negative rates are also equal.
Equal opportunity is defined in terms
of a preferred outcome.
When $y=0$ is preferred, it requires that
$\pr(\hat y=1\giv y=0,\bsx_p)=\pr(\hat y=1\giv y=0)$
That is, the false positive rate
is unaffected by the protected variables.

\cite{chou:2017} considers a score $s$
(such as $\hat y$) to be well calibrated if
$\pr(y=1\giv s ,\bsx_p)=\pr(y=1\giv s)$ for all levels of $s$.
A score attains predictive parity if
$\pr(y=1\giv s>s_*,\bsx_p)=\pr(y=1\giv s>s_*)$ for all thresholds $s_*$.
The distinction between well calibratedness
and predictive parity is relevant when
there are more than two levels for
the score.

There is some debate about when or whether
using protected variables can lead to
improved fairness.
See \cite{xian:2020} who gives a summary of
legal issues surrounding fairness
and \cite{corb:etal:2017}
who study whether
imposing calibration or other criteria
might adversely affect the groups they
are meant to help.

\subsection{Aggregation and disaggregation}

The additivity axiom of Shapley value
is convenient for us.  It means that
importances for the residual $y_i-\hat y_i$
is simply the difference of importances
for $y_i$ and $\hat y_i$.
We can aggregate importances from subjects
to the whole data set by summing, or more
interpretably averaging
over $t=1,\dots,n$.
We can also disaggregate from
the whole data set to subsets of
special interest by averaging cohort
Shapley values over target subjects
$t\in v\subset 1{:}n$.

\section{COMPAS data}\label{sec:compas}

COMPAS is a tool from Northpointe Inc.
for judging the risk that a criminal defendant will
commit another crime (re-offend)
within two years. Each subject is rated into one of
ten deciles with higher deciles considered higher
risk of reoffending.
\cite{angw:2016} investigated whether that algorithm
was biased against Black people. They obtained data
for subjects in Broward County Florida,
including the COMPAS decile, the subjects' race,
age, gender, number of prior crimes and whether the
crime for which they were charged is a felony
or not.  \cite{angw:2016} describe how they
processed their data including how they found
followup data on offences committed and how
they matched those to subjects
for whom they had prior COMPAS scores.
They also note that race was not one of the variables
used in the COMPAS predictions.

The example is controversial.
\cite{angw:2016} find that COMPAS
is biased because it gave a
higher rate of false positives for Black
subjects and a higher rate of false negatives
for White subjects.
\cite{flores2016false} and \cite{dieterich2016compas} disagree raising the issue of $\hat y\giv y$
fairness versus $y\giv \hat y$ fairness.
The prevelance of reoffences differed
between Black and White subjects
forcing $y\giv\hat y$ and $\hat y \giv y$
notions to be incompatible.
\cite{flores2016false} also questioned whether the
subjects in the data set were at comparable
stages in the legal process to those for
which COMPAS was designed.

Following \cite{chou:2017} we focus on just Black
and White subjects.  That provides a sample of
5278 subjects from among the original 6172 subjects.
As in that paper we record the number of prior
convictions as a categorical variable with
five levels: 0, 1--3, 4--6, 7--10 and $>$10.
Following \cite{angw:2016}, we record the
subjects' ages as a categorical variable
with three levels:
$<$25, 25--45, $>$45.
Also, following \cite{chou:2017}, we consider the
prediction $\hat y_i$ to be $1$ if subject
$i$ is in deciles 5--10 and $\hat y_i=0$
for subject $i$ in deciles 1--4.

\section{Exploration of the COMPAS data}\label{sec:dataanalysis}

In this section we use cohort Shapley
to study some fairness issues in the COMPAS
data, especially individual level metrics.  
We have selected what we found to be
the strongest and most interesting findings.
The larger set of figures and tables from which
these are drawn are in the appendix.
The appendix also compares some conventional
group fairness metrics for Black and White
subjects.

We have computed Shapley impacts for these
responses: $y_i$, $\hat y_i$, $y_i-\hat y_i$,
$\fp_i = 1\{ y_i=0\,\&\, \hat y_i=1\}$ and
$\fn_i = 1\{ y_1=1\,\&\, \hat y_i=1\}$.
If $\fp_i=1$ then subject $i$ received a
false positive prediction. Note that the
sample average value of $\fp_i$
$$\hat\e(\fp_i)=\frac{n_{01}}{n_{\sumdot\sumdot}}
=\fpr\times \frac{n_{0\sumdot}}{n_{\sumdot\sumdot}}
=\fpr\times (1-p).$$
Being wrongly predicted to reoffend is an
adverse outcome that we study.  Its expected
value is the FPR times the fraction of
non-reoffending subjects.
Similarly, $\hat\e(\fn_i)=\fnr\times p$.

\subsection{Graphical analysis}
\begin{figure}[t]
  \centering
  \includegraphics[width=1.0 \textwidth]{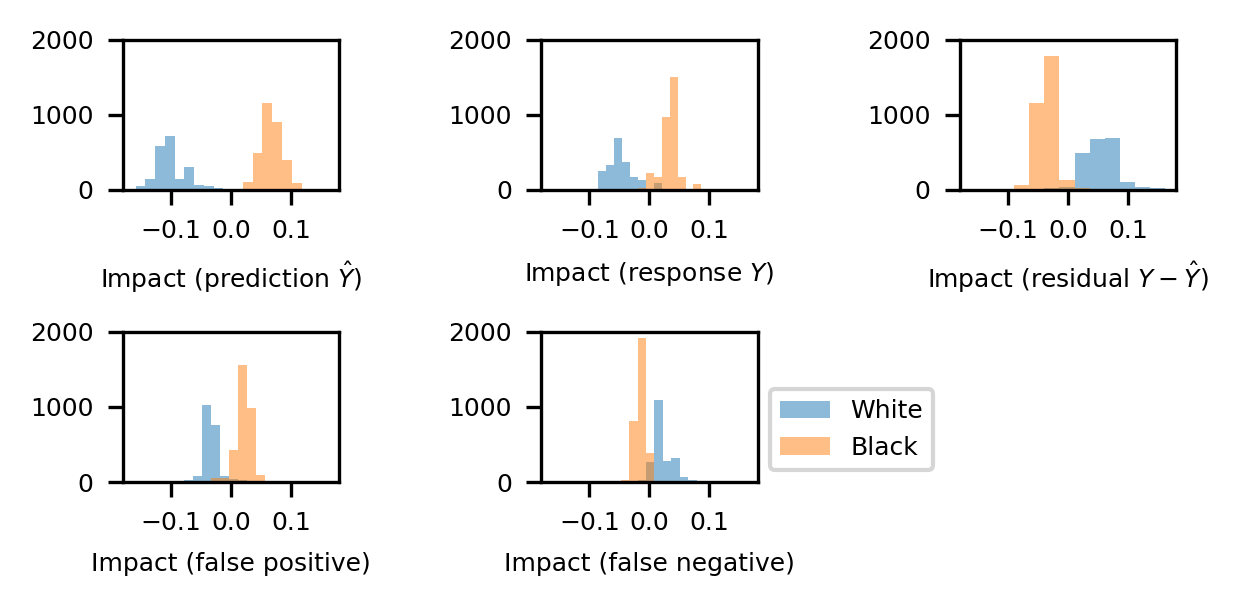}
  \caption{Histogram of cohort Shapley value of race factor on COMPAS recidivism data.}
  \label{fig:compas_cs_race}
\end{figure}

Figure~\ref{fig:compas_cs_race} shows histograms
of Shapley impacts of race for the subjects
in the COMPAS data.  The first panel there reproduces
the data from Figure~\ref{fig:compas_dp_cs_r_hist}
showing a positive impact for every Black subject
and a negative one for every White subject
for the prediction $\hat y$.
For the actual response $y$, the histograms
overlap slightly.  By additivity of Shapley
value the impacts for $y-\hat y$ can be
found by subtracting the impact for $\hat y$
from that for $y$ for each subject $i=1,\dots,n$.
The histograms of $y_i-\hat y_i$ show that
the impact of race on the residual is typically
positive for White subjects and negative
for Black subjects.

Overlap of distributions is seen in equal opportunity (false positive rate: FPR) and false negative rate (FNR). In there, small number of adversely affected White and beneficially
affected Black are observed.
We can inspect further on these overlapped subjects to understand which conditions are exceptional cases.
Also we can inspect on tail of the distribution to understand the extreme cases where bias is very likely observed.

\begin{figure}[t]
  \centering
  \includegraphics[width=1.0 \textwidth]{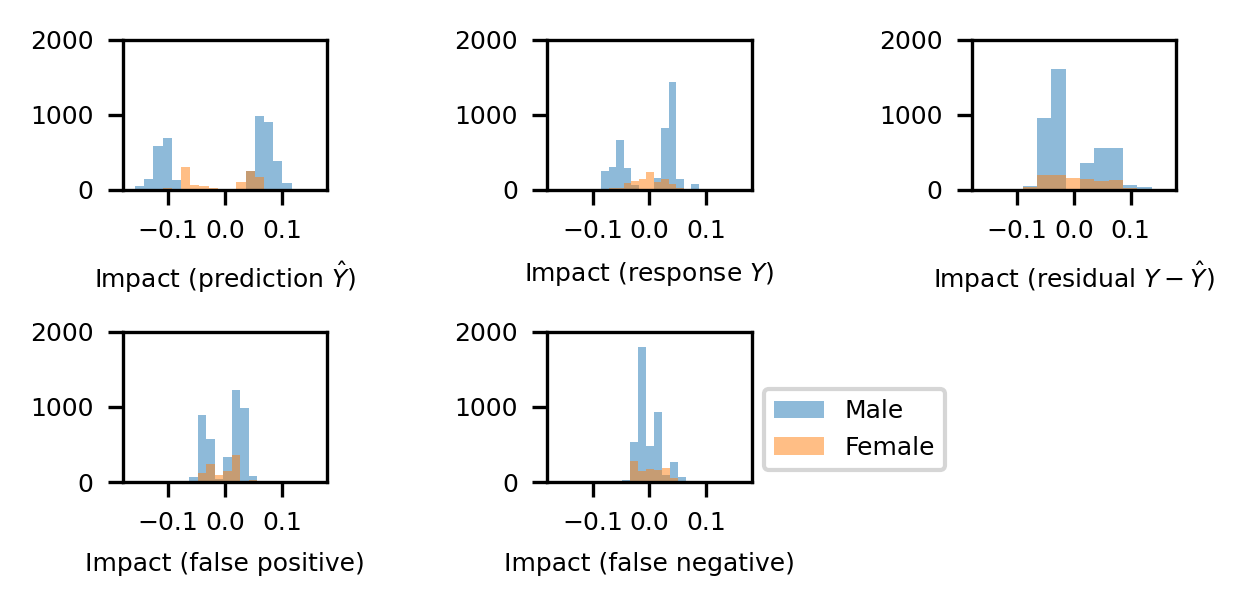}
  \caption{Histogram of cohort Shapley value of race factor for each gender on COMPAS recidivism data.}
  \label{fig:compas_cs_race_gender}
\end{figure}

Figure~\ref{fig:compas_cs_race_gender} shows
histograms of Shapley impacts for race
but now they are color coded by
the subjects' gender.
We see that the impact on $\hat y$ is bimodal
by race for both male and female subjects,
but the effect
is larger in absolute value for male subjects.
The impacts for the response, the residual
and false negative and positive values do not
appear to be bimodal by race for female subjects,
but they do for male subjects.  The case of FN is
perhaps different. The modes in Figure~\ref{fig:compas_cs_race}
are close together and not so apparent in
Figure~\ref{fig:compas_cs_race_gender} which
has slightly different bin boundaries.
It is clear from these figures that the
race differences we see are much stronger
among male subjects.

The appendix includes some figures that each
include 25 histograms.
They are Shapley impact histograms for all five
features color coded by each of the five
features.  There is one such figure for each
of $y$, $\hat y$, $y-\hat y$, $\fp$ and $\fn$.

\subsection{Tabular summary}
Tables~\ref{tab:csi_prediction} through~\ref{tab:csi_fn}
in the appendix record mean Shapley
impacts for the five predictors
and our responses measures, disaggregated
by race and gender.

\begin{table}[t]
\centering
\begin{tabular}{lcccc}
\toprule
Variable & White & Black & Male & Female \\
\midrule
race\_factor &$\phm$0.054 & $-$0.035 & $-$0.001 &$\phm$0.003 \\
gender\_factor & $-$0.001 &$\phm$0.001 &$\phm$0.020 & $-$0.083 \\
\bottomrule\\
\end{tabular}
\smallskip
\begin{tabular}{lcccc}
\toprule
Variable & White-Male & White-Female & Black-Male & Black-Female \\
\midrule
race\_factor &$\phm$0.057 &$\phm$0.042 & $-$0.036 & $-$0.030 \\
gender\_factor &$\phm$0.023 & $-$0.083 &$\phm$0.018 & $-$0.083 \\
\bottomrule\\
\end{tabular}
\caption{\label{tab:csi_residual_subset}
Mean cohort Shapley impact of groups on residual $y - \hat y$.
}
\end{table}

We take a particular interest in the
residual $y_t-\hat y_t$.
It equals $1$ for false negatives and $-1$
for false positives and $0$ when the prediction
was correct.
Table~\ref{tab:csi_residual_subset} shows
a subset of the largest values for
the residual $y_i-\hat y_i$.

What we see there is that revealing that
a subject is Black tells us that, on average,
that subject's residual $y_t-\hat y_t$
is decreased by $3.6$\%.  Revealing that
the subject is White increases the residual
by $5.4$\%. Revealing race makes very little
difference to the residual averaged over
male or over female subjects.
Revealing gender makes quite a large difference
of $-8.3$\% for female subjects and $+2.0\%$
for male subjects.

\begin{figure}[t]
  \centering
  \includegraphics[width=1.0 \textwidth]{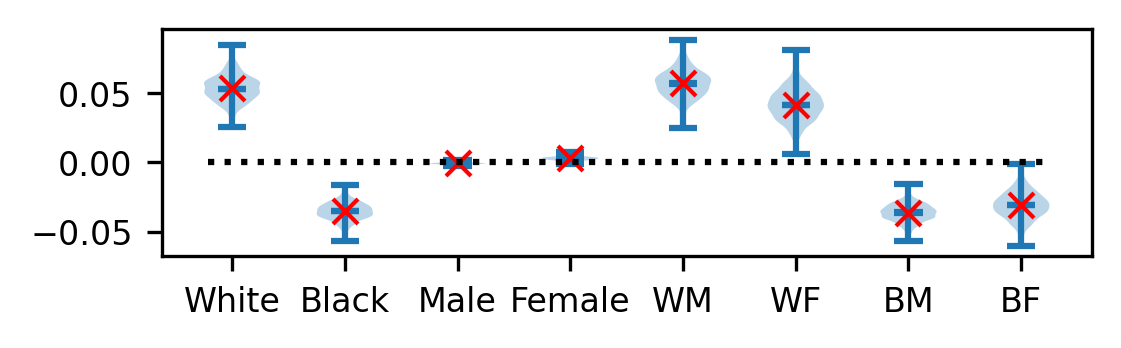}
  \caption{Bayesian bootstrap violin plot of aggregated cohort Shapley race factor impact of groups on residual $y - \hat y$. 
  Red crosses represent aggregated cohort Shapley values of groups without bootstrap.}
  \label{fig:agg_race_impact_violin}
\end{figure}

To judge the uncertainty in the values in
Table~\ref{tab:csi_residual_subset}
we applied the Bayesian bootstrap of \cite{rubi:1981}. 
That algorithm randomly reweights each data point 
by a unit mean exponential random variable. By 
never fully deleting any observation it allows 
one to also bootstrap individual subjects' Shapley 
values though there is not space to do so in this
article. Figure~\ref{fig:agg_race_impact_violin} shows
violin plots of 1000 bootstrapped cohort Shapley values.

\subsection{FPR and FNR revisited}

The risk of being falsely predicted
to reoffend involves two factors:
having $y_i=0$ and having $\hat y_i=1$
given that $y_i=0$. FPR is commonly
computed only over subjects with $y_i=0$.
Accordingly, in this section we study
it by subsetting the subjects to
$\{ i\mid y_i=0\}$ and finding cohort Shapley
value of $\hat y_i=1$ for the features.

\begin{figure}[t]
  \centering
  \includegraphics[width=1.0 \textwidth]{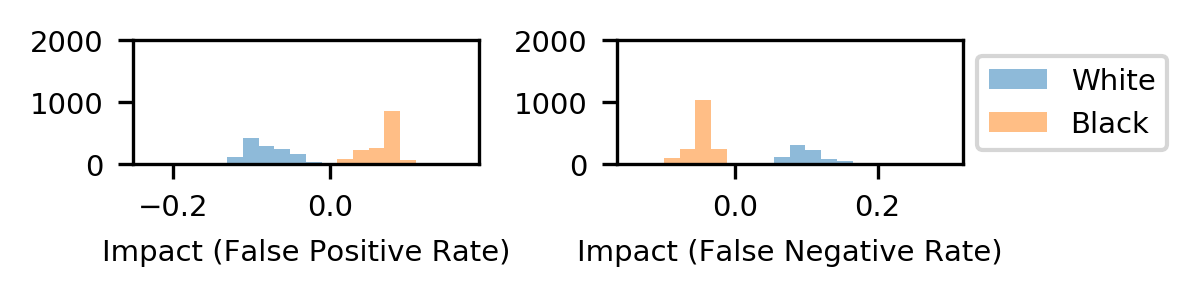}
  \caption{Histogram of cohort Shapley value of race factor on COMPAS recidivism data for false positive subsetting to $y_i = 0$ and false negative subsetting to $y_i = 1$.}
  \label{fig:compas_cs_fprfnr}
\end{figure}

Figure~\ref{fig:compas_cs_fprfnr} shows cohort Shapley impact of race factor conditioned on race factor for FPR and FNR.
The distribution of impacts conditioned on race are separated in the figure.
This conditional analysis shows a stronger disadvantage for Black subjects.

\section{Conclusions}\label{sec:conclusions}

Variable importance in statistics and machine
learning ordinarily considers three issues:
whether changing $x_j$ has a causal future
impact on $y$, whether omitting $x_j$ from
a data set degrades prediction accuracy
and whether changing $x_j$ has an
important mechanical effect on $f(\bsx)$
\citep{jian:owen:2003}. These are all
different from each other and the third
choice is the most useful one for
explaining the decisions of a
prediction method.

Cohort Shapley \citep{mase:owen:seil:2019}
looks at a fourth issue: whether
learning the value
of $x_{tj}$ in a sample is
informative about the prediction
$\hat y_t$.  It is well suited to
settings such
as the COMPAS data from Broward County,
where: the original algorithm is
not available to researchers, the
protected
variable of greatest interest was not
included in the model,
and the set of subjects for which fairness
is of interest is different from the set on
which the algorithm was trained.

Cohort Shapley does not address counterfactuals,
such as whether subject $t$ would
have had $\hat y_t=0$, but for the fact
that $x_{tj}=B$ instead of $W$.
At face value, such counterfactuals directly
address the issue of interest. Unfortunately
there can be many other variables and combinations
of variables that would also have changed
the outcome, providing an equally plausible
explanation. Taking account of them
all combinations can bring in
implausible or even
impossible variable combinations.

Cohort Shapley avoids using
unobserved combinations of variables.
In a fairness setting we might find
that every combination of variables is
logically and physically possible,
though some combinations
may still be quite unlikely, such as the
largest number of prior offences at
the youngest of ages.

We do not offer any conclusions on the
fairness of the COMPAS algorithm.
There is a problem of missing variables
that requires input from domain
experts.

If some
such variable is not in our data set
then accounting for it could change
the magnitude or even the sign of
some of the effects we see.
Some other issues that we believe
require input from domain experts
are: choosing the appropriate set of
subjects to study, determining which
responses are relevant to fairness
and which combination of protected
and unprotected variables are most
suitable to include.
We do however
believe that given a set of subjects
and a set of variables, that cohort
Shapley supports many different
graphical and numerical investigations.
It can do so for variables not used
in a black box at hand, and it does
not even require access to the black
box function.

Cohort Shapley value can be used to detect algorithmic
bias in an algorithm circumventing some
limitations described above.
As with any testing method, there is the possibility
of false positives and false negatives.
Any bias that is detected must be followed up by
further examination.
The positive social outcomes of our method
arise from providing a tool that can detect
and illustrate biases that might have otherwise been missed.
Negative social outcomes can also arise from false
positives and false negatives due to missing variables.

\section*{Acknowledgments}
This work was supported by
the US National Science Foundation under grant IIS-1837931
and by a grant from Hitachi, Ltd.

\bibliographystyle{apalike}
\bibliography{paper}

\begin{thebibliography}{}

\bibitem[Adler et~al., 2018]{adle:2018}
Adler, P., Falk, C., Friedler, S.~A., Nix, T., Rybeck, G., Scheidegger, C.,
  Smith, B., and Venkatasubramanian, S. (2018).
\newblock Auditing black-box models for indirect influence.
\newblock {\em Knowledge and Information Systems}, 54(1):95--122.

\bibitem[Angwin et~al., 2016]{angw:2016}
Angwin, J., Larson, J., Mattu, S., and Kirchner, L. (2016).
\newblock Machine bias: there’s software used across the country to predict
  future criminals. and it’s biased against blacks.

\bibitem[Berk et~al., 2018]{berk:2018}
Berk, R., Heidari, H., Jabbari, S., Kearns, M., and Roth, A. (2018).
\newblock Fairness in criminal justice risk assessments: The state of the art.
\newblock {\em Sociological Methods \& Research}.

\bibitem[Chouldechova, 2017]{chou:2017}
Chouldechova, A. (2017).
\newblock Fair prediction with disparate impact: A study of bias in recidivism
  prediction instruments.
\newblock {\em Big data}, 5(2):153--163.

\bibitem[Chouldechova and Roth, 2018]{chou:roth:2018}
Chouldechova, A. and Roth, A. (2018).
\newblock The frontiers of fairness in machine learning.
\newblock Technical report, arXiv:1810.08810.

\bibitem[Corbett-Davies and Goel, 2018]{corb:goel:2018}
Corbett-Davies, S. and Goel, S. (2018).
\newblock The measure and mismeasure of fairness: A critical review of fair
  machine learning.
\newblock {\em arXiv preprint arXiv:1808.00023}.

\bibitem[Corbett-Davies et~al., 2017]{corb:etal:2017}
Corbett-Davies, S., Pierson, E., Feller, A., Goel, S., and Huq, A. (2017).
\newblock Algorithmic decision making and the cost of fairness.
\newblock In {\em Proceedings of the 23rd acm sigkdd international conference
  on knowledge discovery and data mining}, pages 797--806.

\bibitem[Datta et~al., 2016]{datt:2016}
Datta, A., Sen, S., and Zick, Y. (2016).
\newblock Algorithmic transparency via quantitative input influence: Theory and
  experiments with learning systems.
\newblock In {\em 2016 IEEE symposium on security and privacy (SP)}, pages
  598--617. IEEE.

\bibitem[Dieterich et~al., 2016]{dieterich2016compas}
Dieterich, W., Mendoza, C., and Brennan, T. (2016).
\newblock {COMPAS} risk scales: Demonstrating accuracy equity and predictive
  parity.
\newblock Technical report, Northpoint Inc.

\bibitem[Flores et~al., 2016]{flores2016false}
Flores, A.~W., Bechtel, K., and Lowenkamp, C.~T. (2016).
\newblock False positives, false negatives, and false analyses: A rejoinder to
  machine bias: There's software used across the country to predict future
  criminals. and it's biased against blacks.
\newblock {\em Federal Probation}, 80(2):38--46.

\bibitem[Friedler et~al., 2019]{frie:etal:2019}
Friedler, S.~A., Scheidegger, C., Venkatasubramanian, S., Choudhary, S.,
  Hamilton, E.~P., and Roth, D. (2019).
\newblock A comparative study of fairness-enhancing interventions in machine
  learning.
\newblock In {\em Proceedings of the conference on fairness, accountability,
  and transparency}, pages 329--338.

\bibitem[Hardt et~al., 2016]{hard:2016}
Hardt, M., Price, E., and Srebro, N. (2016).
\newblock Equality of opportunity in supervised learning.
\newblock In {\em Thirtieth Conference on Neural Information Processing Systems
  (NIPS 2016)}.

\bibitem[Holland, 1988]{holl:1988}
Holland, P.~W. (1988).
\newblock Causal inference, path analysis and recursive structural equations
  models.
\newblock Technical Report 88--81, ETS Research Report Series.

\bibitem[Hooker and Mentch, 2019]{hook:ment:2019:tr}
Hooker, G. and Mentch, L. (2019).
\newblock Please stop permuting features: An explanation and alternatives.
\newblock Technical report, arXiv:1905.03151.

\bibitem[Jiang and Owen, 2003]{jian:owen:2003}
Jiang, T. and Owen, A.~B. (2003).
\newblock Quasi-regression with shrinkage.
\newblock {\em Mathematics and Computers in Simulation}, 62(3-6):231--241.

\bibitem[Kleinberg et~al., 2016]{klei:mull:ragh:2016}
Kleinberg, J., Mullainathan, S., and Raghavan, M. (2016).
\newblock Inherent trade-offs in the fair determination of risk scores.
\newblock Technical report, arXiv:1609.05807.

\bibitem[Kumar et~al., 2020]{kuma:2020}
Kumar, I.~E., Venkatasubramanian, S., Scheidegger, C., and Friedler, S. (2020).
\newblock Problems with {Shapley}-value-based explanations as feature
  importance measures.
\newblock In {\em The 37th International Conference on Machine Learning (ICML
  2020)}.

\bibitem[Lundberg and Lee, 2017]{lund:lee:2017}
Lundberg, S.~M. and Lee, S.-I. (2017).
\newblock A unified approach to interpreting model predictions.
\newblock In {\em Advances in Neural Information Processing Systems}, pages
  4765--4774.

\bibitem[Mase et~al., 2019]{mase:owen:seil:2019}
Mase, M., Owen, A.~B., and Seiler, B.~B. (2019).
\newblock Explaining black box decisions by {Shapley} cohort refinement.
\newblock Technical report, arXiv:1911.00467.

\bibitem[Mill, 1843]{mill:1843}
Mill, J.~S. (1843).
\newblock {\em A system of logic}.
\newblock Parker, London.

\bibitem[Molnar, 2018]{moln:2018}
Molnar, C. (2018).
\newblock {\em Interpretable machine learning: A Guide for Making Black Box
  Models Explainable}.
\newblock Leanpub.

\bibitem[Razavi et~al., 2021]{raza:etal:2021}
Razavi, S., Jakeman, A., Saltelli, A., Prieur, C., Iooss, B., Borgonovo, E.,
  Plischke, E., Piano, S.~L., Iwanaga, T., Becker, W., Tarantola, S.,
  Guillaume, J. H.~A., Jakeman, J., Gupta, H., Milillo, N., Rabitti, G.,
  Chabridon, V., Duan, Q., Sun, X., Smith, S., Sheikholeslami, R., Hosseini,
  N., Asadzadeh, M., Puy, A., Kucherenko, S., and Maier, Holger, R. (2021).
\newblock The future of sensitivity analysis: An essential discipline for
  systems modeling and policy support.
\newblock {\em Environmental Modelling \& Software}, 137:104954.

\bibitem[Ribeiro et~al., 2016]{ribe:etal:2016}
Ribeiro, M.~T., Singh, S., and Guestrin, C. (2016).
\newblock Why should {I} trust you?: Explaining the predictions of any
  classifier.
\newblock In {\em Proceedings of the 22nd ACM SIGKDD international conference
  on knowledge discovery and data mining}, pages 1135--1144, New York. ACM.

\bibitem[Rubin, 1981]{rubi:1981}
Rubin, D.~B. (1981).
\newblock The {Bayesian} bootstrap.
\newblock {\em The annals of statistics}, 9(1):130--134.

\bibitem[Saltelli et~al., 2008]{salt:etal:2008}
Saltelli, A., Ratto, M., Andres, T., Campolongo, F., Cariboni, J., Gatelli, D.,
  Saisana, M., and Tarantola, S. (2008).
\newblock {\em Global sensitivity analysis: the primer}.
\newblock John Wiley \& Sons.

\bibitem[Shapley, 1953]{shap:1953}
Shapley, L.~S. (1953).
\newblock A value for n-person games.
\newblock In Kuhn, H.~W. and Tucker, A.~W., editors, {\em Contribution to the
  Theory of Games II (Annals of Mathematics Studies 28)}, pages 307--317.
  Princeton University Press, Princeton, NJ.

\bibitem[Sobol', 1993]{sobo:1993}
Sobol', I.~M. (1993).
\newblock Sensitivity estimates for nonlinear mathematical models.
\newblock {\em Mathematical Modeling and Computational Experiment}, 1:407--414.

\bibitem[Sundararajan and Najmi, 2020]{sund:najm:2019}
Sundararajan, M. and Najmi, A. (2020).
\newblock The many {Shapley} values for model explanation.
\newblock In {\em The 37th International Conference on Machine Learning (ICML
  2020)}.

\bibitem[Wei et~al., 2015]{wei:lu:song:2015}
Wei, P., Lu, Z., and Song, J. (2015).
\newblock Variable importance analysis: A comprehensive review.
\newblock {\em Reliability Engineering \& System Safety}, 142:399--432.

\bibitem[Xiang, 2020]{xian:2020}
Xiang, A. (2020).
\newblock Reconciling legal and technical approaches to algorithmic bias.
\newblock {\em Tennessee Law Review}, 88(3):2021.

\end{thebibliography}

\newpage
\appendix
\section{Appendix}
This appendix includes conventional group fairness metrics and whole bias distribution analysis results by cohort Shapley on COMPAS data example.

\begin{figure}[t]
  \centering
  \includegraphics[width=1.0 \textwidth]{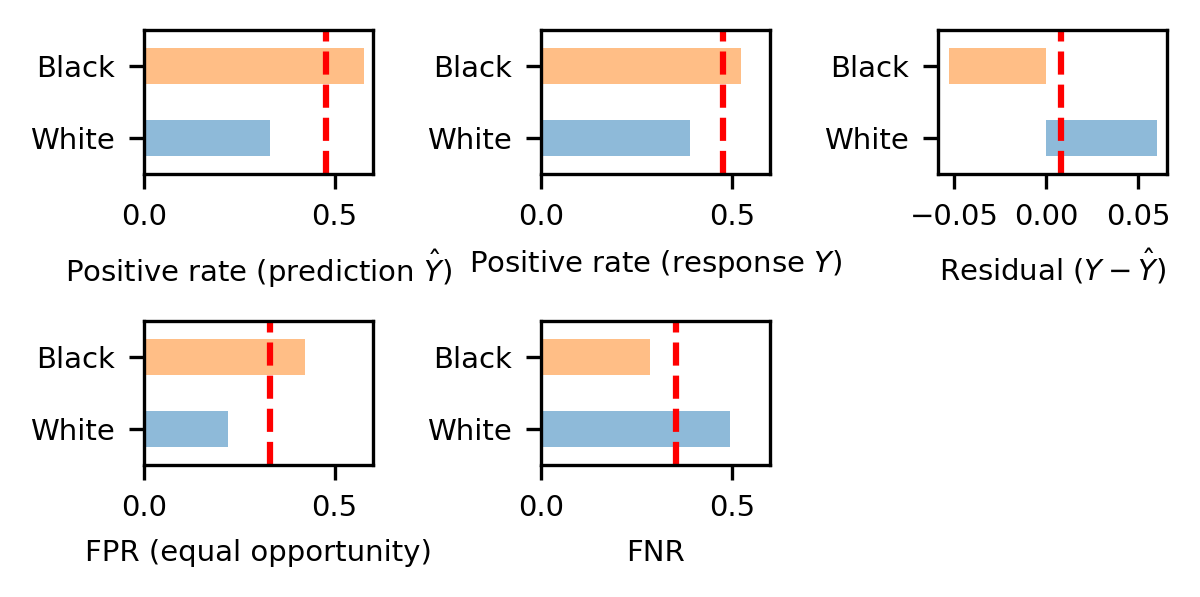}
  \caption{Group fairness metrics on COMPAS recidivism data.}
  \label{fig:compas_gf}
\end{figure}

Figure~\ref{fig:compas_gf} shows some
conventional group
fairness metrics for the COMPAS data set.
Horizontal bars show group specific means
and the vertical dashed lines show population
means.
We see that Black subjects had a higher
average value of $\hat y$ than White subjects.
Black subjects also had a higher average of $y$ but
a lower average residual $y-\hat y$.
Using $B$ and $W$ to denote the two
racial groups,
$\hat\e(y-\hat y\giv B)\doteq -0.035$
and $\hat\e(y-\hat y\giv W)\doteq 0.054$.
The FPR was higher for Black subjects and the FNR
was higher for White subjects.

\begin{figure*}[t]
  \centering
  \includegraphics[width=1.0 \textwidth]{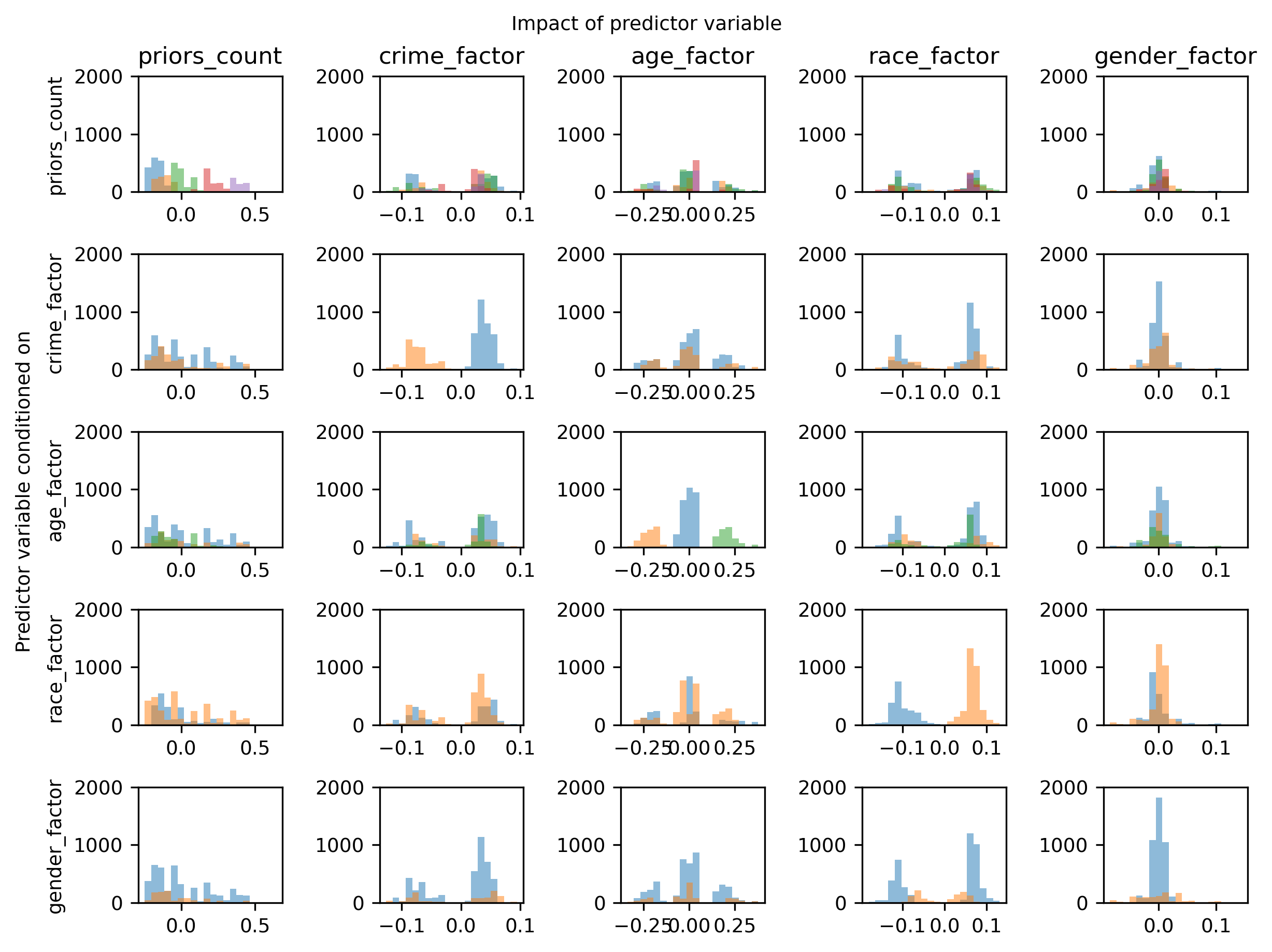}
  \caption{Histograms of cohort Shapley value indicating individualized bias for demographic parity of prediction on a variable for each category in a conditioned variable.}
  \label{fig:compas_dp_cs_hist}
\end{figure*}

\begin{figure*}[t]
  \centering
  \includegraphics[width=1.0 \textwidth]{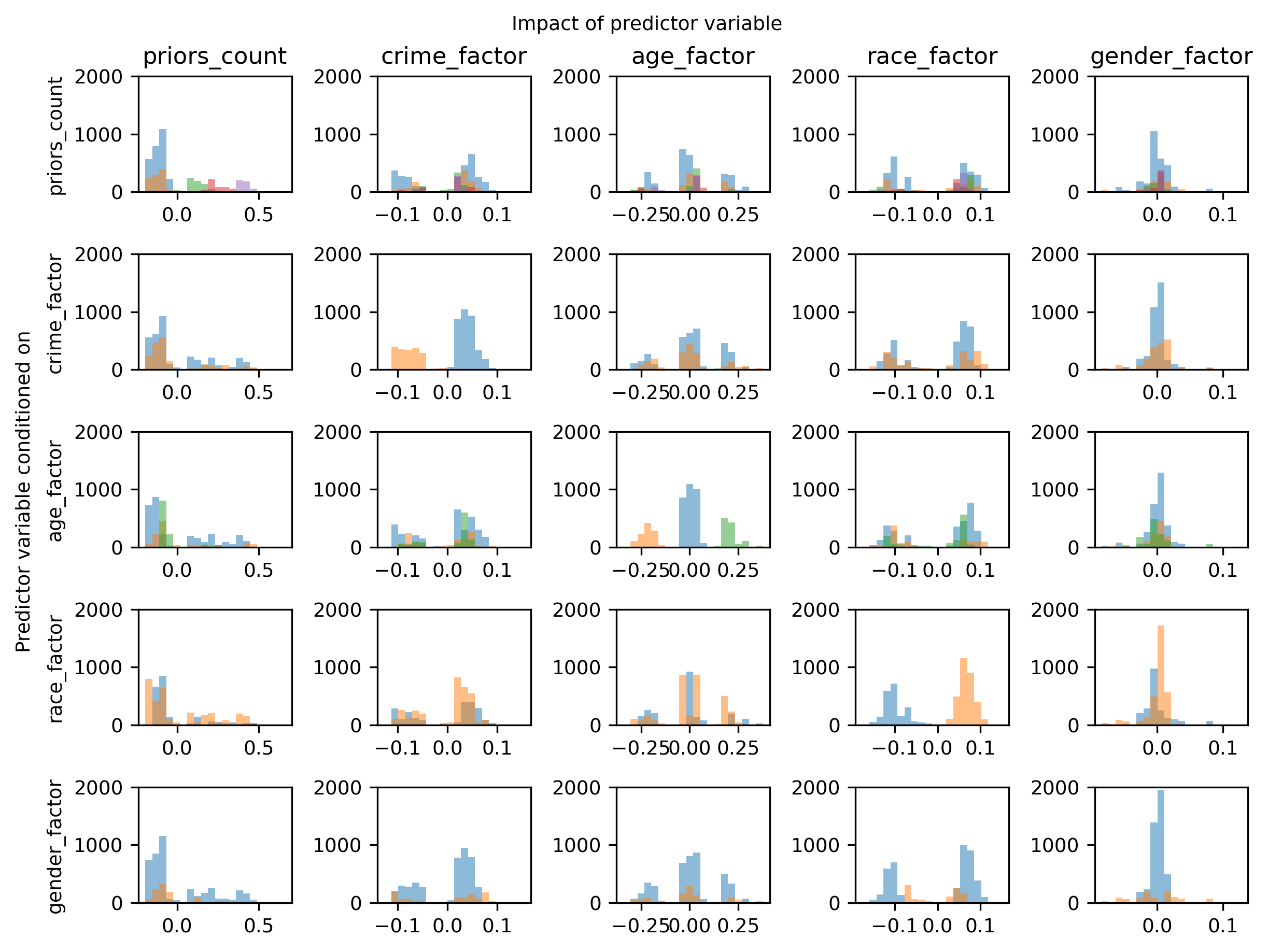}
  \caption{Histograms of cohort Shapley value indicating individualized bias for demographic parity of response on a variable for each category in a conditioned variable.}
  \label{fig:compas_y_cs_hist}
\end{figure*}

\begin{figure*}[t]
  \centering
  \includegraphics[width=1.0 \textwidth]{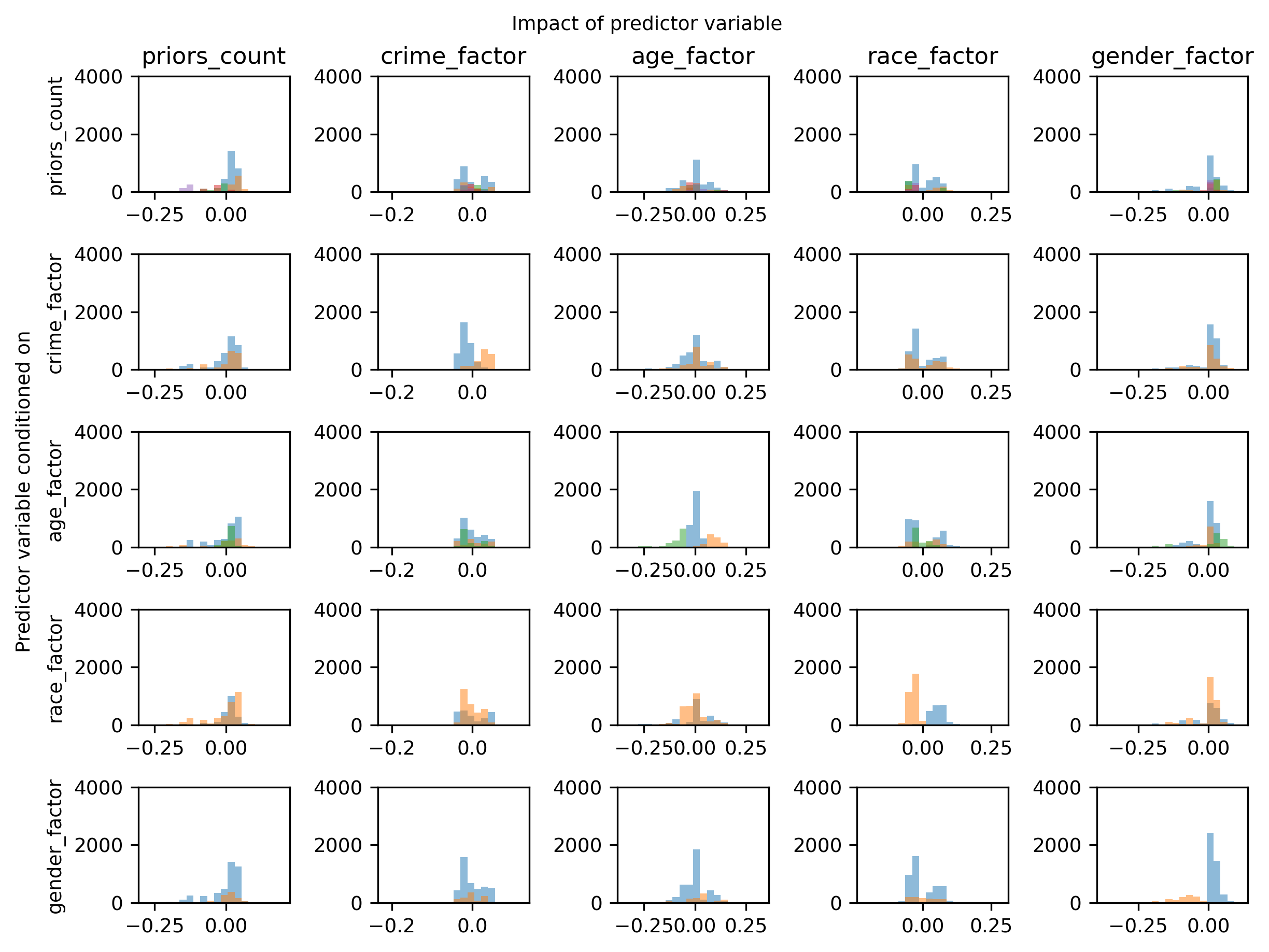}
  \caption{Histograms of cohort Shapley value indicating individualized bias for residual on a variable for each category in a conditioned variable.}
  \label{fig:compas_res_cs_hist}
\end{figure*}

\begin{figure*}[t]
  \centering
  \includegraphics[width=1.0 \textwidth]{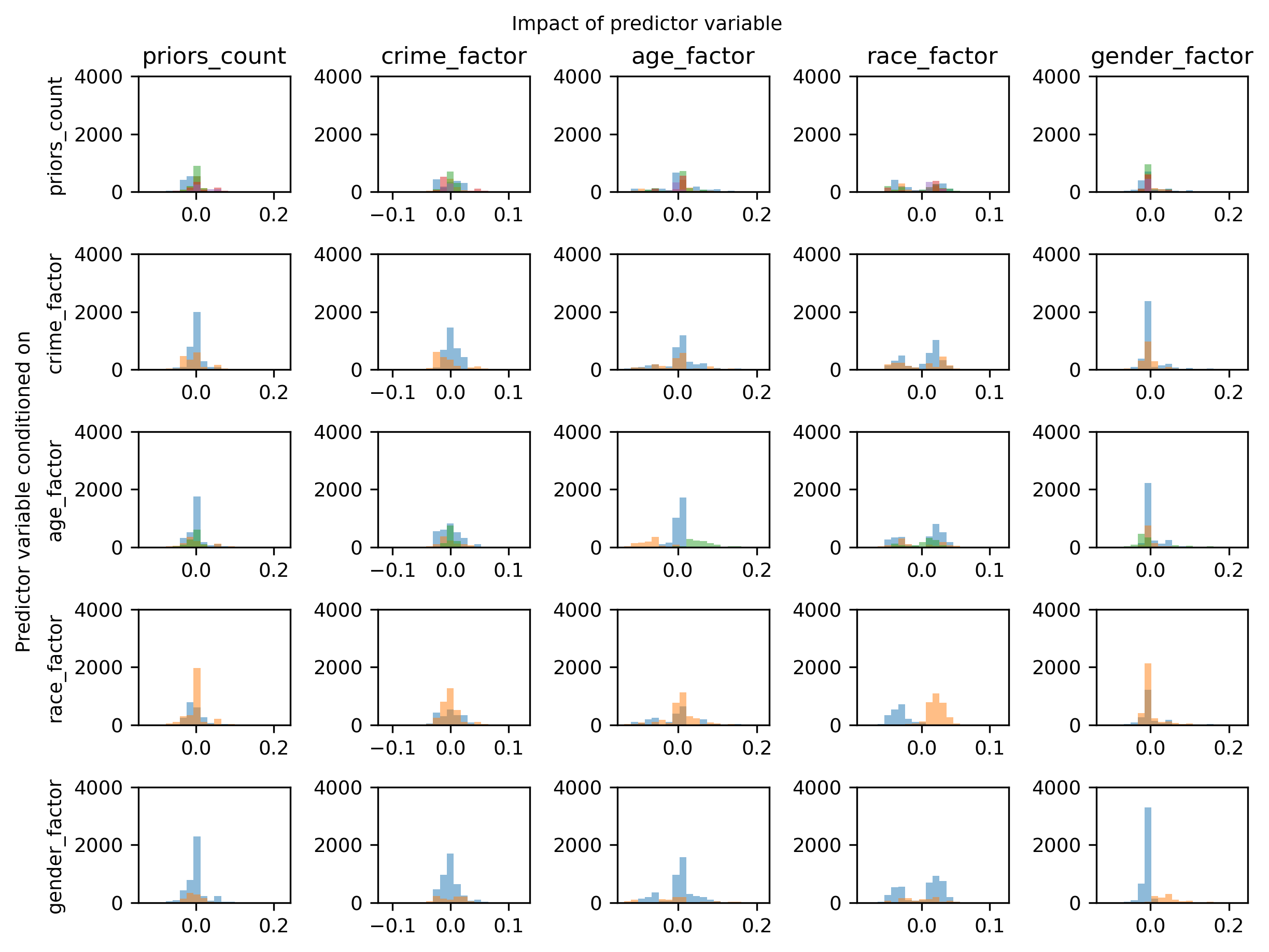}
  \caption{Histograms of cohort Shapley value indicating individualized bias for equal opportunity (false positive rate) on a variable for each category in a conditioned variable.}
  \label{fig:compas_eo_cs_hist}
\end{figure*}

Figure~\ref{fig:compas_dp_cs_hist} describes 5 x 5 matrix of histograms on individualized bias (impact)
on demographic parity of prediction. The columns represent the variables where we look into their biases
and the rows represent the conditioned variables for grouping when we generate the histogram.
The colors in the figure indicate categories in the conditioned variables.
Figure~\ref{fig:compas_y_cs_hist} describes histograms of individualized bias on demographic parity of response.
Figure~\ref{fig:compas_res_cs_hist} describes histograms of individualized bias on residual.
Figure~\ref{fig:compas_eo_cs_hist} describes histograms of individualized bias on false positive that corresponds to equal opportunity.
Figure~\ref{fig:compas_eo_cs_hist} describes histograms of individualized bias on false negative.

\begin{figure*}[t]
  \centering
  \includegraphics[width=1.0 \textwidth]{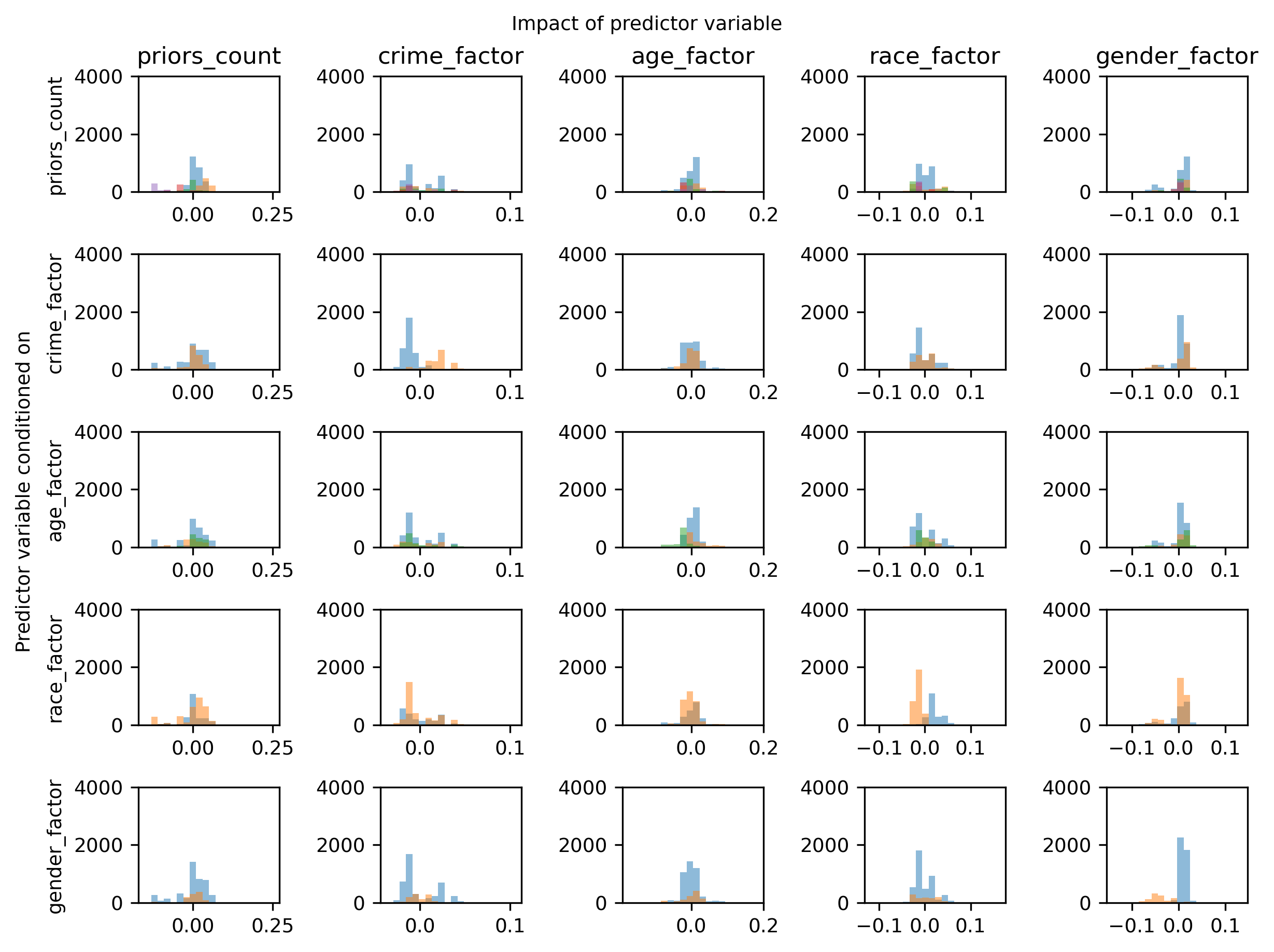}
  \caption{Histograms of cohort Shapley value indicating individualized bias for false negative rate on a variable for each category in a conditioned variable.}
  \label{fig:compas_fnr_cs_hist}
\end{figure*}

\begin{table}\centering
\caption{\label{tab:csi_prediction}
Mean cohort Shapley impact of groups on prediction $\hat y$.
}
\smallskip
\begin{tabular}{lcccc}
\toprule
Variable & White & Black & Male & Female \\
\midrule
priors\_count & $-$0.024 &$\phm$0.016 &$\phm$0.006 & $-$0.023 \\
crime\_factor & $-$0.004 &$\phm$0.003 &$\phm$0.001 & $-$0.004 \\
age\_factor & $-$0.018 &$\phm$0.012 &$\phm$0.000 & $-$0.001 \\
race\_factor & $-$0.101 &$\phm$0.067 &$\phm$0.002 & $-$0.006 \\
gender\_factor & $-$0.001 &$\phm$0.000 &$\phm$0.000 & $-$0.002 \\
\bottomrule\\
\end{tabular}
\begin{tabular}{lcccc}
\toprule
Variable & White-Male & White-Female & Black-Male & Black-Female \\
\midrule
priors\_count & $-$0.023 & $-$0.027 &$\phm$0.024 & $-$0.019 \\
crime\_factor & $-$0.002 & $-$0.010 &$\phm$0.003 &$\phm$0.001 \\
age\_factor & $-$0.017 & $-$0.020 &$\phm$0.011 &$\phm$0.015 \\
race\_factor & $-$0.112 & $-$0.065 &$\phm$0.072 &$\phm$0.045 \\
gender\_factor & $-$0.008 &$\phm$0.025 &$\phm$0.006 & $-$0.025 \\
\bottomrule\\
\end{tabular}
\end{table}

\begin{table}\centering
\caption{\label{tab:csi_response}
Mean cohort Shapley impact of groups on response $y$.
}
\smallskip
\begin{tabular}{lcccc}
\toprule
Variable & White & Black & Male & Female \\
\midrule
priors\_count & $-$0.018 &$\phm$0.012 &$\phm$0.004 & $-$0.018 \\
crime\_factor & $-$0.003 &$\phm$0.002 &$\phm$0.001 & $-$0.003 \\
age\_factor & $-$0.010 &$\phm$0.007 &$\phm$0.000 & $-$0.001 \\
race\_factor & $-$0.048 &$\phm$0.032 &$\phm$0.001 & $-$0.003 \\
gender\_factor & $-$0.002 &$\phm$0.001 &$\phm$0.021 & $-$0.085 \\
\bottomrule\\
\end{tabular}
\smallskip
\begin{tabular}{lcccc}
\toprule
Variable & White-Male & White-Female & Black-Male & Black-Female \\
\midrule
priors\_count & $-$0.017 & $-$0.021 &$\phm$0.017 & $-$0.015 \\
crime\_factor & $-$0.001 & $-$0.006 &$\phm$0.002 &$\phm$0.000 \\
age\_factor & $-$0.010 & $-$0.009 &$\phm$0.007 &$\phm$0.007 \\
race\_factor & $-$0.055 & $-$0.023 &$\phm$0.035 &$\phm$0.014 \\
gender\_factor &$\phm$0.015 & $-$0.058 &$\phm$0.024 & $-$0.107 \\
\bottomrule\\
\end{tabular}
\end{table}

\begin{table}\centering
\caption{\label{tab:csi_residual}
Mean cohort Shapley impact of groups on residual $y - \hat y$.
}
\smallskip
\begin{tabular}{lcccc}
\toprule
Variable & White & Black & Male & Female \\
\midrule
priors\_count &$\phm$0.007 & $-$0.004 & $-$0.001 &$\phm$0.006 \\
crime\_factor &$\phm$0.001 & $-$0.001 & $-$0.000 &$\phm$0.001 \\
age\_factor &$\phm$0.008 & $-$0.005 & $-$0.000 &$\phm$0.000 \\
race\_factor &$\phm$0.054 & $-$0.035 & $-$0.001 &$\phm$0.003 \\
gender\_factor & $-$0.001 &$\phm$0.001 &$\phm$0.020 & $-$0.083 \\
\bottomrule\\
\end{tabular}
\smallskip
\begin{tabular}{lcccc}
\toprule
Variable & White-Male & White-Female & Black-Male & Black-Female \\
\midrule
priors\_count &$\phm$0.007 &$\phm$0.007 & $-$0.006 &$\phm$0.005 \\
crime\_factor &$\phm$0.001 &$\phm$0.003 & $-$0.001 & $-$0.000 \\
age\_factor &$\phm$0.007 &$\phm$0.011 & $-$0.004 & $-$0.009 \\
race\_factor &$\phm$0.057 &$\phm$0.042 & $-$0.036 & $-$0.031 \\
gender\_factor &$\phm$0.023 & $-$0.083 &$\phm$0.018 & $-$0.083 \\
\bottomrule\\
\end{tabular}
\end{table}

\begin{table}\centering
\caption{\label{tab:csi_fp}
Mean cohort Shapley impact of groups on false positive.
}
\smallskip
\begin{tabular}{lcccc}
\toprule
Variable & White & Black & Male & Female \\
\midrule
priors\_count & $-$0.002 &$\phm$0.002 &$\phm$0.000 & $-$0.002 \\
crime\_factor & $-$0.001 &$\phm$0.000 &$\phm$0.000 & $-$0.001 \\
age\_factor & $-$0.006 &$\phm$0.004 &$\phm$0.000 & $-$0.000 \\
race\_factor & $-$0.033 &$\phm$0.022 &$\phm$0.001 & $-$0.002 \\
gender\_factor &$\phm$0.001 & $-$0.000 & $-$0.011 &$\phm$0.044 \\
\bottomrule\\
\end{tabular}
\smallskip
\begin{tabular}{lcccc}
\toprule
Variable & White-Male & White-Female & Black-Male & Black-Female \\
\midrule
priors\_count & $-$0.002 & $-$0.003 &$\phm$0.002 & $-$0.001 \\
crime\_factor & $-$0.000 & $-$0.002 &$\phm$0.000 &$\phm$0.001 \\
age\_factor & $-$0.005 & $-$0.008 &$\phm$0.003 &$\phm$0.007 \\
race\_factor & $-$0.034 & $-$0.022 &$\phm$0.023 &$\phm$0.015 \\
gender\_factor & $-$0.013 &$\phm$0.047 & $-$0.009 &$\phm$0.042 \\
\bottomrule\\
\end{tabular}
\end{table}

\begin{table}\centering
\caption{\label{tab:csi_fn}
Mean cohort Shapley impact of groups on false negative.
}
\smallskip
\begin{tabular}{lcccc}
\toprule
Variable & White & Black & Male & Female \\
\midrule
priors\_count &$\phm$0.004 & $-$0.003 & $-$0.001 &$\phm$0.004 \\
crime\_factor &$\phm$0.001 & $-$0.001 & $-$0.000 &$\phm$0.001 \\
age\_factor &$\phm$0.002 & $-$0.001 & $-$0.000 &$\phm$0.000 \\
race\_factor &$\phm$0.021 & $-$0.014 & $-$0.000 &$\phm$0.001 \\
gender\_factor & $-$0.001 &$\phm$0.000 &$\phm$0.010 & $-$0.039 \\
\bottomrule\\
\end{tabular}
\smallskip
\begin{tabular}{lcccc}
\toprule
Variable & White-Male & White-Female & Black-Male & Black-Female \\
\midrule
priors\_count &$\phm$0.005 &$\phm$0.004 & $-$0.004 &$\phm$0.004 \\
crime\_factor &$\phm$0.001 &$\phm$0.002 & $-$0.001 &$\phm$0.000 \\
age\_factor &$\phm$0.001 &$\phm$0.002 & $-$0.001 & $-$0.002 \\
race\_factor &$\phm$0.021 &$\phm$0.020 & $-$0.014 & $-$0.015 \\
gender\_factor &$\phm$0.010 & $-$0.037 &$\phm$0.009 & $-$0.041 \\
\bottomrule\\
\end{tabular}
\end{table}

\end{document}